\documentclass[10pt,twocolumn,letterpaper]{article}

\usepackage{cvpr}     
\usepackage{graphicx}
\usepackage{booktabs}
\definecolor{cvprblue}{rgb}{0.21,0.49,0.74}
\usepackage[pagebackref,breaklinks,colorlinks,allcolors=cvprblue]{hyperref}

\def\paperID{*****} % *** Enter the Paper ID here
\def\confName{CVPR}
\def\confYear{2026}

\title{NudgeVAD: Language-Nudged End-to-End Driving via FiLM Residuals}

\author{
Chieh-Chi Yang$^{*}$ \quad Yu-Hsiang Chen$^{*}$ \quad Yi-Ting Chen\\
National Yang Ming Chiao Tung University\\
$^{*}$Equal contribution
}
\begin{document}
\maketitle
\begin{abstract}
Natural-language instructions promise controllable end-to-end driving, but their
benefit can be hidden when planners already receive reliable high-level
commands. We propose \textbf{NudgeVAD}, a frozen-planner residual framework that
uses language as a calibrated \emph{nudge} to a VAD trajectory. With
identity-initialized FiLM and a zero-initialized residual head, NudgeVAD is
equivalent to the frozen planner at initialization, so learned deviations arise
only from language-conditioned residuals.

We evaluate NudgeVAD along a \emph{command-reliability axis}. With reliable
commands, language improves the initial planner but becomes nearly redundant
once compared against \textsc{VAD-FT (Uncond)}, a compute-matched VAD model
fine-tuned without language. With random commands, however, language becomes
essential: detaching text degrades $\mathrm{ADE}_{6s}$ to $3.166$ m, while
NudgeVAD with text recovers $\mathbf{2.806}$ m and outperforms
\textsc{VAD-FT (Uncond)} by $0.312$ m. These results show that language is not
universally additive; it is most valuable when the categorical command channel
is unreliable.
\end{abstract}    
\section{Introduction}

\begin{figure*}[t]
    \centering
    \includegraphics[width=0.95\textwidth]{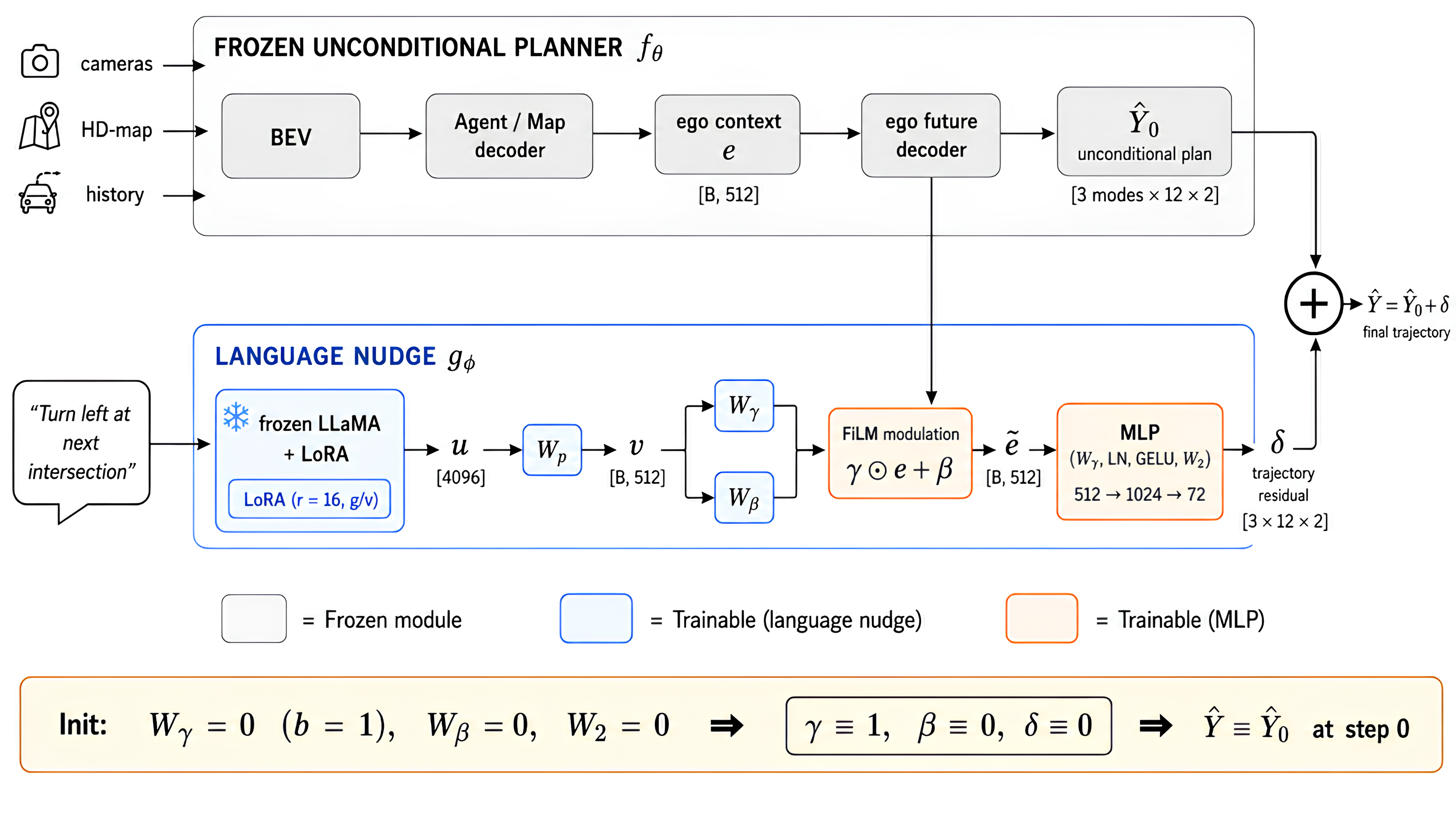}
    \caption{
    Overview of \textbf{NudgeVAD}. A frozen VAD planner predicts an
    unconditional trajectory $\hat{\mathbf{Y}}_0$ from sensor-map context,
    ego history, and a command $\mathbf{c}$ inferred from past-only lanelet
    geometry. The instruction is encoded by a frozen language model with
    lightweight adapters, FiLM-modulates the planner ego feature $\mathbf{e}$,
    and predicts a residual $\boldsymbol{\Delta}$. The final output is
    $\hat{\mathbf{Y}}=\hat{\mathbf{Y}}_0+\boldsymbol{\Delta}$. Identity
    initialization makes $\boldsymbol{\Delta}=0$ at step zero, so NudgeVAD is
    initially identical to the frozen planner.
    }
    \label{fig:architecture}
\end{figure*}

End-to-end autonomous driving has made strong progress with BEV-based and
query-based planners that encode sensor-map context, reason over agents and
lanes, and regress future ego trajectories~\cite{bevformer,uniad,vad,vadv2}.
However, these planners typically infer the ego future from history and scene
context alone, treating the ego vehicle like any other traffic participant.
This ignores a key distinction: the ego vehicle may have access to privileged
intent, such as a route command or a human instruction. Natural language offers
a flexible way to express this intent, e.g., \textit{``turn left at the next
intersection''} or \textit{``follow the white truck''}. The doScenes benchmark
therefore asks whether language can improve trajectory prediction beyond a
strong driving planner~\cite{doscenes,nuscenes}.

Surprisingly, language is not always helpful in practice. The public
instruction-conditioned doScenes result reports a negative conditioning gain,
and we observe similar failures when naively fine-tuning a large language
encoder with a strong planner. We argue that the issue is not merely model
capacity or fusion design, but a structural redundancy between language and the
categorical command channel. In VAD-style planners, the future decoder predicts
multiple maneuver-conditioned trajectories, and a command selects the left,
right, or straight mode~\cite{vad}. When this command is reliable, it already
provides much of the maneuver information that language would otherwise add.
As a result, language can improve an initial checkpoint, yet become nearly
redundant once compared against a compute-matched unconditional VAD model
fine-tuned without language.

We propose \textbf{NudgeVAD}, a frozen-planner residual framework for studying
when language is actually useful. Instead of jointly fine-tuning the planner,
NudgeVAD keeps the VAD planner frozen and uses language only to predict a
trajectory residual. A frozen LLaMA encoder maps the instruction to a sentence
embedding, which FiLM-modulates the planner ego feature~\cite{film}; a
lightweight MLP then predicts a residual added to the VAD trajectory. With
identity-initialized FiLM and a zero-initialized residual head, NudgeVAD is
exactly equivalent to the frozen planner at initialization, so any later
deviation is learned through the language residual.

Our key finding is that the value of language depends on command reliability.
With reliable commands, NudgeVAD substantially improves the initial planner, but
adding the same language adapter on top of \textsc{VAD-FT (Uncond)}---a
compute-matched VAD model fine-tuned without language---yields only
$+0.003$ m ADE gain. When commands are unreliable, the conclusion reverses:
detaching text degrades $\mathrm{ADE}$ to $3.166$ m, while NudgeVAD with text
recovers $\mathbf{2.806}$ m and outperforms \textsc{VAD-FT (Uncond)} by
$0.312$ m. Language is therefore not universally additive; it becomes most
valuable when the categorical command channel is unreliable.

\paragraph{Contributions.}
We make three contributions:
\begin{itemize}
    \item We identify command-language redundancy as a key reason why
    instruction conditioning may fail to improve strong VAD-style planners.
    \item We introduce \textbf{NudgeVAD}, a frozen-planner residual framework
    with identity initialization that isolates language-conditioned trajectory
    corrections without perturbing the pretrained planner.
    \item We evaluate language conditioning along a command-reliability axis
    and show that, under unreliable commands, NudgeVAD outperforms a
    compute-matched unconditional VAD fine-tuning baseline by $0.312$ m.
\end{itemize}
\section{Method}
\label{sec:method}

We propose \textbf{NudgeVAD}, a frozen-planner residual framework for
instruction-conditioned driving. Given a pretrained VAD planner, NudgeVAD keeps
the planner fixed and trains only a lightweight language branch to predict a
trajectory correction. This isolates language from planner fine-tuning and
preserves the planner's learned geometric priors.

A key protocol detail is that we do \emph{not} use ground-truth commands.
VAD-style planners often select the planning mode using a future-derived
command or oracle navigation input~\cite{vad,sparsedrive,ssr}. Under doScenes,
however, the command must be available at inference and must not leak future
ego motion. We therefore infer $\mathbf{c}$ only from past ego history and
HD-map lanelet geometry.

\subsection{Residual Instruction Conditioning}
\label{sec:formulation}

Given ego history $\mathbf{H}$, sensor-map context $\mathbf{S}$, command
$\mathbf{c}\in\{0,1\}^{3}$, and instruction $\mathbf{t}$, the task is to
predict the future ego trajectory $\mathbf{Y}\in\mathbb{R}^{T_f\times2}$.
We use $T_f=12$ waypoints, corresponding to $6$ seconds at $2$ Hz. As shown in
Fig.~\ref{fig:architecture}, NudgeVAD decomposes prediction into a frozen VAD
trajectory and a language residual:
\begin{equation}
\label{eq:residual_formulation}
\hat{\mathbf{Y}}
=
\underbrace{f_{\theta}(\mathbf{H},\mathbf{S},\mathbf{c})}_{\text{frozen VAD}}
+
\underbrace{g_{\phi}(\mathbf{e},\mathbf{t})}_{\text{language residual}},
\end{equation}
where $\mathbf{e}$ is the VAD ego planning feature. During NudgeVAD training,
$\theta$ is fixed and only the residual branch $g_{\phi}$ is optimized.

\subsection{Frozen 3-Mode VAD Planner}
\label{sec:frozen_planner}

We instantiate $f_{\theta}$ with VAD-Tiny~\cite{vad}. Given camera and map
inputs, VAD encodes the scene into an ego planning feature
$\mathbf{e}\in\mathbb{R}^{D_e}$ and predicts three command-conditioned
trajectories:
\begin{equation}
\label{eq:uncond_output}
\hat{\mathbf{Y}}_0
=
\mathrm{EgoFutDec}_{\theta}(\mathbf{e})
\in\mathbb{R}^{3\times T_f\times2}.
\end{equation}
The command selects the final trajectory,
$\hat{\mathbf{y}}_0=\hat{\mathbf{Y}}_0[\mathbf{c}]$. This routing structure is
central to our analysis: when $\mathbf{c}$ is reliable, it already carries the
maneuver class and can make language redundant; when $\mathbf{c}$ is
unreliable, the instruction becomes the remaining maneuver-level signal.

\subsection{Language Nudge}
\label{sec:language_residual}

The residual branch encodes the instruction with a frozen LLaMA encoder and
trainable LoRA adapters~\cite{lora}. Given final token states
$\mathbf{H}_t\in\mathbb{R}^{L\times D_t}$ and attention mask $\mathbf{m}$, we
form a sentence embedding by masked mean pooling and project it to the planner
feature space:
\begin{equation}
\label{eq:text_pooling}
\mathbf{u}
=
\frac{1}{\sum_{\ell=1}^{L}m_{\ell}}
\sum_{\ell=1}^{L}m_{\ell}\mathbf{H}_t[\ell],
\qquad
\mathbf{v}=W_p\mathbf{u}.
\end{equation}
The projected instruction feature generates FiLM parameters~\cite{film} for the
ego feature:
\begin{equation}
\label{eq:film}
\boldsymbol{\gamma}=W_{\gamma}\mathbf{v}+\mathbf{1},
\qquad
\boldsymbol{\beta}=W_{\beta}\mathbf{v},
\qquad
\tilde{\mathbf{e}}
=
\boldsymbol{\gamma}\odot\mathbf{e}
+
\boldsymbol{\beta}.
\end{equation}
A lightweight MLP then predicts a command-conditioned residual:
\begin{equation}
\label{eq:residual_head}
\boldsymbol{\Delta}
=
W_2\,\mathrm{GELU}
\left(
\mathrm{LN}(W_1\tilde{\mathbf{e}})
\right)
\in\mathbb{R}^{3\times T_f\times2}.
\end{equation}
The final prediction is
\begin{equation}
\label{eq:final_prediction}
\hat{\mathbf{Y}}
=
\hat{\mathbf{Y}}_0+\boldsymbol{\Delta},
\qquad
\hat{\mathbf{y}}
=
\hat{\mathbf{Y}}[\mathbf{c}].
\end{equation}
Thus, language does not replace the planner; it nudges the VAD trajectory
through an additive residual.

\subsection{Identity Initialization}
\label{sec:identity_init}

To prevent the language branch from perturbing the frozen planner before
learning, we initialize
\begin{equation}
\resizebox{0.7\linewidth}{!}{$
\begin{aligned}
W_{\gamma}&=0,\quad
W_{\beta}=0,\quad
b_{\gamma}=\mathbf{1},\quad
b_{\beta}=0,\quad
W_2=0,\\
\boldsymbol{\gamma}&=\mathbf{1},\quad
\boldsymbol{\beta}=\mathbf{0},\quad
\boldsymbol{\Delta}=\mathbf{0}
\quad\Rightarrow\quad
\hat{\mathbf{Y}}=\hat{\mathbf{Y}}_0 .
\end{aligned}
$}
\label{eq:init_invariance}
\end{equation}
Therefore, NudgeVAD is exactly equivalent to the frozen VAD planner at
initialization. Any later deviation is learned through the language residual.

\subsection{Command-Reliability Probe}
\label{sec:cmd_reliability}

To separate language from the categorical command channel, we evaluate two
command regimes. Let $\mathbf{c}_{\mathrm{lanelet}}$ be the past-only lanelet
command. We define
\begin{equation}
\resizebox{0.7\linewidth}{!}{$
\mathbf{c}
=
\mathcal{N}_{\rho}(\mathbf{c}_{\mathrm{lanelet}})
=
\begin{cases}
\mathbf{c}_{\mathrm{lanelet}}, & \rho=\mathrm{reliable},\\
\mathrm{OneHot}(\mathrm{Uniform}\{0,1,2\}), & \rho=\mathrm{random}.
\end{cases}
$}
\label{eq:cmd_neutral}
\end{equation}
The reliable regime tests whether language adds information beyond a strong
command channel. The random regime removes this categorical shortcut at both
training and inference, forcing maneuver information to enter through language.
In both regimes, with-text and no-text passes use the same $\mathbf{c}$, so
$\Delta\mathrm{ADE}$ isolates the contribution of the language residual.

\subsection{Training Objective}
\label{sec:training}

NudgeVAD trains only the language residual branch. Given ground truth
$\mathbf{Y}$, we minimize the weighted L1 trajectory loss
\begin{equation}
\label{eq:loss}
\mathcal{L}_{\mathrm{traj}}
=
\sum_{t=1}^{T_f}
w_t
\left\|
\hat{\mathbf{y}}_t-\mathbf{Y}_t
\right\|_1
+
\lambda_{\mathrm{end}}
\left\|
\hat{\mathbf{y}}_{T_f}-\mathbf{Y}_{T_f}
\right\|_1 .
\end{equation}
No auxiliary language loss is used. We use three stabilizing choices: removing
the scalar residual gate, increasing residual-MLP capacity, and keeping frozen
perception modules in evaluation mode to prevent batch-normalization drift.
\section{Experiments}
\label{sec:exp}

We evaluate whether language provides information beyond a categorical command
channel. We distinguish two baselines: \textsc{VAD-Init}, the initial VAD
checkpoint used by NudgeVAD, and \textsc{VAD-FT (Uncond)}, a compute-matched
VAD model fine-tuned on the same data without language. This distinction is
important: language may improve an initial planner, yet become redundant once
the unconditional planner is further optimized.

\subsection{Setup}
\label{sec:exp-setup}

We evaluate on doScenes~\cite{doscenes}, built on nuScenes~\cite{nuscenes}.
All methods predict one open-loop $6$\,s ego trajectory with $T_f=12$ waypoints
at $2$\,Hz. We report ADE, FDE, and
\[
\Delta\mathrm{ADE}
=
\mathrm{ADE}_{\mathrm{no\text{-}text}}
-
\mathrm{ADE}_{\mathrm{with\text{-}text}},
\]
computed using the same model and the same command $\mathbf{c}$.

We compare two command regimes. In the \emph{reliable} regime,
$\mathbf{c}$ is the past-only lanelet command. In the \emph{random} regime,
$\mathbf{c}$ is replaced by a uniformly sampled one-hot command at both
training and inference. To control for optimization budget,
\textsc{VAD-FT (Uncond)} continues training the VAD trunk for the same
additional $60$ epochs as NudgeVAD, but without language.

\paragraph{Stop override.}
During evaluation, we apply a conservative stop override only when both text
and motion indicate stopping: a hard-stop cue, no conflicting action or
stop-related noun phrase, word count $\leq 12$, and history speed
$\leq 2\,\mathrm{m/s}$.

\subsection{Command Reliability Determines the Value of Language}
\label{sec:exp-main}

\begin{table}[t]
\centering
\small
\caption{
Command-reliability comparison. Under reliable commands, language improves
\textsc{VAD-Init} but becomes nearly redundant on top of
\textsc{VAD-FT (Uncond)}. Under random commands, language provides a large
same-model gain and outperforms the unconditional baseline.
}
\label{tab:main_cmd}
\resizebox{\linewidth}{!}{%
\begin{tabular}{llcccc}
\toprule
Cmd & Method & ADE $\downarrow$ & FDE $\downarrow$ &
$\Delta$ADE $\uparrow$ & Gain over \textsc{VAD-FT} \\
\midrule
Reliable
& \textsc{VAD-Init}                         & 3.146& 7.238& --- & --- \\
& NudgeVAD on \textsc{VAD-Init}             & \textbf{2.957}& \textbf{6.763} & $+0.293$& $-0.228$\\
& \textsc{VAD-FT (Uncond)}                  & 2.729& 6.396& --- & --- \\
& NudgeVAD on \textsc{VAD-FT}               & 2.726& 6.376& $+0.003$& $+0.003$\\
\midrule
Random
& \textsc{VAD-Stage1}                       & 3.590& 7.333& --- & --- \\
& \textsc{VAD-FT (Uncond)}                  & 3.118& 6.517& --- & --- \\
& NudgeVAD w/o text                         & 3.166& 6.722& --- & --- \\
& \textbf{NudgeVAD w/ text}                 & \textbf{2.806}& \textbf{6.148}&
$\mathbf{+0.36}$& $\mathbf{+0.312}$\\
\bottomrule
\end{tabular}%
}
\end{table}

Table~\ref{tab:main_cmd} shows two regimes. With reliable commands, NudgeVAD
substantially improves the initial VAD checkpoint, reducing ADE from $3.146$ to
$2.957$. However, once the unconditional planner is compute-matched through
additional fine-tuning, adding the same language adapter gives only
$+0.003$ m ADE gain. Thus, reliable commands make language useful on a weaker
initial planner, but nearly redundant on top of a stronger unconditional one.

With random commands, the conclusion changes sharply. The no-text NudgeVAD pass
fails because the command no longer provides a reliable maneuver signal, while
the with-text model recovers strong performance, reducing ADE from $3.166$ to
$2.086$. It also outperforms \textsc{VAD-FT (Uncond)} by $0.312$ m. This
supports our central claim: language is most valuable when the categorical
command channel is unreliable.

\subsection{What Makes the Adapter Work?}
\label{sec:exp-ablation}

\begin{table}[t]
\centering
\small
\setlength{\tabcolsep}{3.5pt}
\caption{
Adapter progression under unreliable command routing. A plain text residual is
not enough; residual capacity and FiLM are the main drivers of the language
gain. The conservative stop override has only a marginal effect.
}
\label{tab:adapter_progression}
\resizebox{\linewidth}{!}{%
\begin{tabular}{lcccccccc}
\toprule
Method & a@1s & a@2s & a@3s & a@4s & a@5s & a@6s & FDE & Gain@6s \\
\midrule
\textsc{VAD-Stage1} no-cmd base
& 0.360& 1.100& 1.605& 2.275& 2.923& 3.590& 7.333& --- \\
\textsc{VAD-FT (Uncond)}
& \textbf{0.337}& 1.088& 1.561& 2.031& 2.526& 3.118& 6.517& $+0.472$\\
Plain text residual
& 0.377& 1.008& 1.589& 2.066& 2.577& 3.170& 6.547& $+0.420$\\
Large residual MLP
& 0.369& 0.912& 1.418& 1.863& 2.367& 2.961& 6.341& $+0.629$\\
\textbf{NudgeVAD (FiLM)}
& 0.351& 0.846& 1.305& 1.731& 2.221& 2.806& 6.148& $+0.784$\\
Stop override
& 0.348& \textbf{0.836}& \textbf{1.290}& \textbf{1.713}& \textbf{2.197}& \textbf{2.774}& \textbf{6.071}& $\mathbf{+0.816}$\\
\bottomrule
\end{tabular}%
}
\end{table}

Table~\ref{tab:adapter_progression} explains why the random-command regime
benefits from language. A plain text residual does not outperform the
compute-matched unconditional baseline at $6$ s. Scaling the residual MLP
closes much of the gap, and FiLM provides the largest improvement, reducing
a@6s to $2.806$ and outperforming \textsc{VAD-FT (Uncond)} by $0.312$ m. The
stop override changes a@6s by only $0.032$ m, indicating that the gain comes
from the learned language residual rather than post-processing.

\paragraph{Limitations.}
ADE is not a complete measure of instruction following. In supplementary
analysis, we find that lower ADE does not always improve semantic compliance,
suggesting that future instructed-driving benchmarks should report both
trajectory error and instruction-compliance metrics.

\paragraph{Implementation.}
NudgeVAD trains for $60$ epochs with AdamW and a cosine learning rate of
$10^{-4}$. The planner is VAD-Tiny~\cite{vad}. \textsc{VAD-FT (Uncond)} uses
the same training budget as NudgeVAD but contains no language branch. The
language encoder is LLaMA-7B with LoRA adapters. No auxiliary language loss is
used.
\section{Conclusion}

We presented \textbf{NudgeVAD}, a frozen-planner residual framework for
instruction-conditioned driving. Our study shows that the value of language is
conditional rather than universal. With reliable commands, language can improve
an initial planner but becomes nearly redundant once compared against a
compute-matched unconditional VAD fine-tuning baseline. With unreliable
commands, however, language provides maneuver intent that the unconditional
planner cannot recover, outperforming \textsc{VAD-FT (Uncond)} by $0.312$ m.

These findings suggest that instruction-conditioned driving should not be
evaluated only under oracle or near-oracle command channels. Future benchmarks
should explicitly vary command reliability and measure not only trajectory
accuracy, but also whether the predicted motion satisfies the instruction.
NudgeVAD provides a simple baseline for studying when language is redundant and
when it is genuinely necessary for planning.
{
    \small
    \bibliographystyle{ieeenat_fullname}
    \bibliography{main}
}


\end{document}